\documentclass{article}
\usepackage{ijcai21}

\usepackage{times}

\usepackage{soul}
\usepackage{url}
\usepackage[hidelinks]{hyperref}
\usepackage[utf8]{inputenc}
\usepackage[small]{caption}
\usepackage{graphicx}
\usepackage{amsmath,amssymb,amsfonts}
\usepackage{algpseudocode}
\usepackage{algorithm}
\usepackage{booktabs}
\usepackage{subcaption}
\usepackage{booktabs}
\usepackage{pifont}
  
\title{Causal Sensitivity Identification using Generative Learning}

\author{
Soma Bandyopadhyay\textsuperscript{1}\footnote{Contact Author}, Sudeshna Sarkar\textsuperscript{2} \\ 
\textsuperscript{1}TCS Research, TATA Consultancy Services Limited, Kolkata, India \\ 
\texttt{soma.bandyopadhyay@tcs.com}
\and
 \\ 
\textsuperscript{2}Department of Computer Science and Engineering, IIT Kharagpur, India \\ 
\texttt{sudeshna@cse.iitkgp.ac.in}
}
\begin{document}

\maketitle

\begin{abstract}
%In this work, we propose a novel method to identify the causal impact using generative learning and apply on prediction task. %through the intervention and the counterfactual analysis using generative learning.
In this work, we propose a novel generative method to identify the causal impact and apply it to prediction tasks. We conduct causal impact analysis using interventional and counterfactual perspectives. First, applying interventions, we identify features that have a causal influence on the predicted outcome, which we refer to as causally sensitive features, 
and second, applying counterfactuals, we evaluate how changes in the cause affect the effect. Our method exploits the Conditional Variational Autoencoder (CVAE) to identify the causal impact and serve as a generative predictor. We are able to reduce confounding bias by identifying causally sensitive features. We demonstrate the effectiveness of our method by recommending the most likely locations a user will visit next in their spatiotemporal trajectory influenced by the causal relationships among various features. %, achieving competitive performance. We use a large real-world human trajectory dataset, GeoLife~\cite{geolife}, and the benchmark Asia Bayesian network . %In these experiments, we evaluate causal impacts to recommend the most likely places a user will visit in their trajectory, achieving competitive performance.% in our predictions.
Experiments on the large-scale GeoLife~\cite{geolife} dataset and the benchmark Asia Bayesian network validate the ability of our method to identify causal impact and improve predictive performance.
\end{abstract}
\section{Introduction}
%\begin{abstract}
%In this work, we propose a novel generative method to identify \textit{causal impact} and apply it to prediction tasks. We frame causal impact analysis from two complementary perspectives: (i) identifying features that \textit{have a causal effect} on the predicted outcome—referred to as \textit{causally sensitive features}, and (ii) evaluating how changes in these features affect the prediction, incorporating both \textit{interventional} and \textit{counterfactual} views of causality. Our approach leverages a Conditional Variational Autoencoder (CVAE) to model the underlying data-generating process and simulate alternative scenarios, enabling counterfactual prediction and causal effect estimation without requiring an explicit causal graph.  
%By identifying causally sensitive features, our method effectively reduces confounding bias. We demonstrate its applicability in next-location prediction within spatiotemporal human mobility data, where user-specific movement patterns are governed by complex causal relationships among features. We evaluate our method on the large-scale real-world GeoLife dataset and show that our causal analysis enhances interpretability and achieves competitive performance in recommending the most likely next place a user will visit. This work bridges causal reasoning and generative modeling, contributing a task-aware, explainable approach to recommendation and sequential prediction.
%\end{abstract}
Determining causal impact is an important need of causal reasoning to understand the causal path among different features. 

In this work, we aim to identify the causal impact in a prediction task through interventional and counterfactual analysis without assuming any prior causal graph, which we name \textbf{causal sensitivity identification}. We define the objectives of causal sensitivity identification as follows:
\begin{enumerate}
   % \item \textbf{Quantification of causal impact:} Measuring the influence of individual features on model predictions under intervention.
    \item Identifying features that have a causal influence on the prediction outcome, and play a key role in preserving causal relationships ~\cite{SBS}, we refer to these features as \textbf{causally sensitive} features.    
    \item  Assessing the impact of changes in causes on their effects.
      \item Identifying causal paths (e.g., $X \rightarrow Y$) between input features and the prediction target, without assuming any prior knowledge of the causal graph.
            \end{enumerate}
            
We propose a \textbf{generative method} for causal impact analysis using Conditional Variational Autoencoder (CVAE) ~\cite{VAE-T} as generative predictor. 

Causality operates on three main levels of the ladder of causation~\cite{bookofwhy}: association, intervention, and counterfactuals. The lowest level captures statistical associations without causal interpretation. The middle level involves interventions, modeled through the do-calculus, which assigns values randomly to a variable, establishes the direct causal relation between cause and effect, and reduces confounding bias. The highest level represents the counterfactual~\cite{countercausal}, revealing cause and effect relationships and their dependence on counterfactuals~\cite{Pearl-Counter} or alternate situations. For example, could different past location sequences result in the same next location? Does a user's location history influence their next movement?
\begin{itemize} 
\item \textbf{Intervention: Identifying causally sensitive feature}  
\begin{enumerate}
\item% We determine the causally sensitive features ($F_{CS}$). We reduce the confounding bias, where the confounding feature as ($F_{CS}$) influences both the cause and the effect. To identify the causally sensitive features we compare the performance of prediction using identical test data, and using the predictor trained in original (factual) train and intervened train data.
We determine the \textbf{causally sensitive} features ($F_{CS}$) that influence both the cause and the effect, potentially acting as confounders. To detect $F_{CS}$, we compare prediction performance on identical test data using two models, one trained on the original (factual) train data and the other trained on intervened data where candidate features are altered.
\item Identified causally sensitive features are used to \textbf{condition} the generative prediction model to ensure the prediction is guided by a true causal relations.
\end{enumerate}
\item \textbf{Counterfactuals: Assessing impact of cause changes} 
\begin{enumerate}
\item We assess the change in effect when the cause has changed using counterfactual analysis and identify the  \textbf{causal path}. In this scenario, we compare the performance of prediction using counterfactual test data obtaining the counterfactual latent representation, and \textbf{original (factual)} test data using the predictor trained in original (factual) train data. 
\item We obtain the counterfactual predictions / generations to determine outcomes in alternate situations.
\end{enumerate}
\end{itemize}

A summary of our contribution is as follows: 
\begin{enumerate}
\item \textbf{Causal sensitivity identification framework}:
\begin{itemize}
%\item We propose a new method to determine the causally sensitive features using interventional analysis for prediction tasks exploiting CVAE based generative predictor.
%\item We identify causal sensitivity in terms of judging the impact of change in cause on its effect using counterfactuals and identify causal path.
\item We propose a new method to determine causally sensitive features using interventional analysis for prediction tasks, exploiting a CVAE based generative predictor.
 \item We identify causal sensitivity by assessing the impact of changes in causes on their effects using counterfactuals, and use this to identify the underlying causal path.
\end{itemize}
\item \textbf{Causally sensitive recommendation/prediction}: 
\begin{itemize}
\item We evaluate our method on the \textbf{Asia} Bayesian network from the BNLearn repository~\cite{bnlearn}, demonstrating its ability to recover known causal dependencies by identifying causally sensitive features and associated causal paths through interventions and counterfactual analysis.
%\item  We evaluate our method on the \textbf{Asia} Bayesian network \cite{bnlearn} to assess its ability to capture known causal dependencies under interventional and counterfactual scenarios by identifying causally sensitive features and the causal path.
\item We apply the proposed framework to the real-world \textbf{GeoLife}~\cite{geolife} human trajectory dataset, where the identified causally sensitive features (e.g., \texttt{start\_time}) are used to condition next location prediction. We further assess the impact of counterfactual changes in past location sequences on future trajectory predictions.
%    \item Our approach successfully identifies all causal paths relevant to the target variable \texttt{dysp}, demonstrating its effectiveness compared to neural causal discovery baselines such as NOTEARS~\cite{zheng2018dags} and Ke et al.~\cite{ke2020learning}.
%\end{itemize}
\end{itemize}
\end{enumerate}
\vspace{-2mm}
\section{Related Work}
We explore the related works on causality focusing on applications of intervention and counterfactuals using neural models and on the trajectory predictions.
%\begin{itemize}

\textbf{Causal interventions and counterfactuals:}
The use of interventions and counterfactuals to uncover cause-effect relationships has emerged as a key area of research, particularly under partially known or unknown causal structures.
\begin{itemize}
    \item Yang et al.~\cite{Yang2018} propose I-MEC to characterize interventional equivalence classes and reduce ambiguity via soft interventions.
    \item Ke et al.~\cite{ke2020learning} develop a neural approach for causal path discovery under unknown interventions without requiring explicit targets.
    \item Dai et al.~\cite{dai2025selection} address causal discovery under selection bias and latent interventions.
    \item Kung et al.~\cite{count-kung} ensure counterfactual stability by assuming sensitive attributes lack ancestors, enforcing prediction consistency.
\item \texttt{Neural generative models:}
\begin{itemize}
\item VAE~\cite{kingma2019auto} based generative models have been widely adopted for counterfactual reasoning and disentangled causal representation learning.
\begin{itemize}
    \item Louizos et al.~\cite{CEVAE} propose CEVAE, which models noisy proxy variables for unobserved confounders based on Pearl’s backdoor criterion~\cite{Pearl-C}.
    \item Yang et al.~\cite{Yang2021CausalVAE} propose CausalVAE, which integrates a linear structural causal model (SCM) with a VAE for counterfactual generation using known causal structures.\end{itemize}
 \item Kuang et al.~\cite{count1} apply generative adversarial networks (GANs)~\cite{goodfellow2014generative} for counterfactual inference under known causal graphs.
 \end{itemize}
    %\item In~\cite{count-kung}, authors assume sensitive attributes like race and gender to have no ancestors in the graph and demonstrate counterfactual stability under this assumption.
    \item \texttt{Causal structure learning:} Numerous methods aim to recover causal structures from data, often by optimizing a score under acyclicity constraints. Among these, NOTEARS~\cite{NOTEARS} is a widely used method that formulates DAG discovery as a continuous and differentiable optimization problem. It is commonly applied to identify pairwise causal relations and can complement intervention-based approaches. Ke et al.~\cite{ke2020learning} focus on causal discovery under unknown intervention targets, aiming to recover the DAG without explicitly addressing prediction performance or counterfactual inference. %Their method treats intervention targets as latent variables and learns the graph structure by optimizing a neural scoring function.
 %Their approach models intervention targets as latent variables and formulates graph learning as a continuous optimization problem, where a neural scoring function is optimized to uncover structural dependencies.
\end{itemize}
 \textbf{Sequence modeling for trajectory prediction:}
Trajectory prediction based on GPS data is widely studied as a sequence modeling task, especially in mobility applications where past behavior influences future outcomes.
\begin{itemize}
\item LSTM-based models, often combined with attention mechanisms~\cite{attention,causalattention}, effectively capture spatiotemporal dependencies in trajectory prediction.
\item Deep learning has been widely applied to trajectory modeling~\cite{Wang2022}, with surveys and benchmarks in~\cite{Rudenko2020,nezhadettehad2024}.
\item Generative models such as VAEs~\cite{salzmann2020trajectron} support distributional forecasting in frameworks like~\cite{movesim,trajgan}, enabling uncertainty-aware predictions.
%\item Recurrent neural networks such as LSTM, often enhanced with attention mechanisms~\cite{attention,causalattention}, have shown strong performance in capturing spatiotemporal dependencies and highlighting relevant temporal features.
%\item Deep learning models have been widely applied to trajectory data~\cite{Wang2022}, with comprehensive surveys and benchmarks provided in~\cite{Rudenko2020,nezhadettehad2024}.
%\item Generative models such as VAEs and their extensions~\cite{salzmann2020trajectron}  are employed in various state-of-the-art trajectory prediction frameworks, including~\cite{movesim}, ~\cite{trajgan}.
\item \hypertarget{trajcomp}{\textbf{Our emphasis:}} We emphasize the following related works to compare our method against theirs on human trajectory prediction using GPS trajectory data.
%\begin{itemize}
%    \item LSTM: Krishna et al.~\cite{krishna2018} showed that LSTM-based models outperform HMM and hierarchical HMM baselines~\cite{hmm} for activity recognition and duration estimation.
%    \item LSTM \+Attention: Li et al.~\cite{li2020} introduced hierarchical attention mechanisms to improve long-term mobility pattern prediction using LSTM architectures.
%    \item DeepMove: DeepMove~\cite{deepmove} jointly embeds multimodal factors such as time, user ID, and location, enhanced with a historical attention module to improve location transition modeling.
%    \item Ref.~\cite{ref_article1} presents a multi-head self-attention (MHSA) model that integrates high-level spatiotemporal features such as visit time and stay duration for effective location sequence modeling.
% \end{itemize}   
 \begin{itemize}
    \item \textbf{LSTM}: The authors ~\cite{krishna2018} show that LSTM-based models outperform HMM and hierarchical HMM baselines~\cite{hmm} for activity recognition and duration estimation.
    \item \textbf{LSTM + Attention}: The authors ~\cite{li2020} introduce hierarchical attention mechanisms to improve long-term mobility pattern prediction using LSTM architectures.
    \item \textbf{DeepMove}: ~\cite{deepmove} jointly embeds multimodal factors such as time, user ID, and location, enhanced with a historical attention module to improve location transition modeling.
%    \item MHSA: ~\cite{ref_article1} presents a multi-head self-attention (MHSA) model that integrates high-level spatiotemporal features e.g., visit time and stay duration for effective location sequence modeling.
\item \textbf{MHSA}:~\cite{ref_article1} proposes a multi-head self-attention (MHSA) model that leverages spatiotemporal features (e.g., visit time, duration) for location sequence modeling.
 \end{itemize}   
\end{itemize}
Despite their success in sequence learning, these models largely ignore causal reasoning and do not consider how interventions or counterfactual variations in inputs affect trajectory predictions.

\section{Methodology}
%Expanding on the objectives stated in the introduction, we describe our method for identifying causal sensitivity using a CVAE-based generative predictor alongside interventional and counterfactual analysis.
We present the proposed causal sensitivity identification method using a CVAE-based generative model as a generative predictor. Our framework incorporates interventional and counterfactual analysis expanding upon the objectives stated in the introduction. %Building on the objectives outlined in the introduction, we detail our causal sensitivity identification method using a CVAE-based generative predictor with interventional and counterfactual analysis%Causal sensitivity identification comprises two main steps: identifying causally sensitive features and assessing the impact of changes in causes on their effects. Counterfactuals are also used to establish the causal path.\
 \begin{figure*}[!bhtp]
    \centering
    \includegraphics[width=0.88\linewidth]{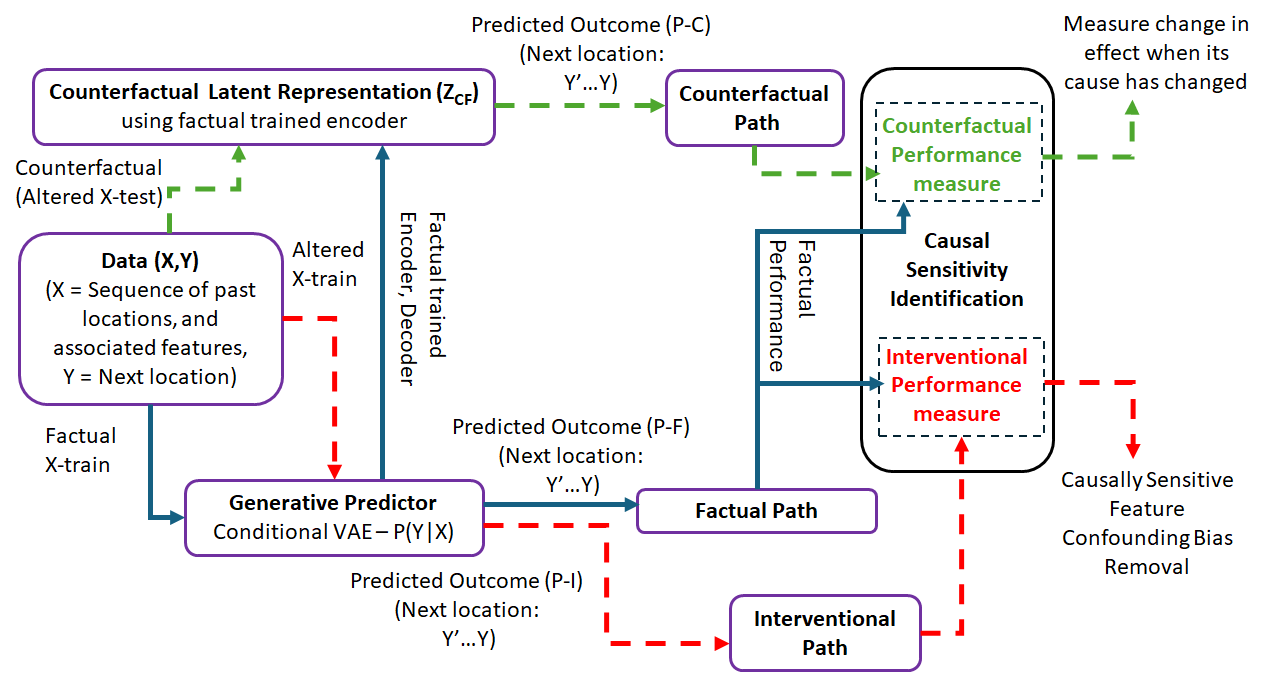}
    \caption{Functional components of the proposed causal sensitivity identification framework with the factual (blue), interventional (red), and counterfactual (green) to evaluate causal influence in prediction tasks.}
    \label{fig1}
\end{figure*}

As illustrated in Figure~\ref{fig1}, the functional components of the proposed method comprises the generative predictor, with the factual path shown is blue solid line, the interventional path shown in red dotted line, counterfactuals path shown in green dotted line along with the counterfactual latent representation $Z_{CF}$, and the input data $D(X,Y)$. 

The proposed framework integrates a generative predictor based on CVAE to evaluate causal sensitivity through three complementary paths: factual, interventional, and counterfactual. The factual path assesses the model’s baseline predictive performance on unaltered data. The interventional path enables the identification of causally sensitive features by measuring performance changes when specific variables are intervened upon (e.g., $do(X_i = x')$). Finally, the counterfactual path estimates the effect of hypothetical changes in the input by generating counterfactual outcomes using the factual latent representations and altered test instances. Together, these components allow for a systematic analysis of causal influence and sensitivity, providing insight into which variables are most critical for accurate prediction and robust decision-making.

The details of the proposed method are described below. 

\textit{Notation:} $X_{\text{train}}$ and $X_{\text{test}}$ denote the factual (unaltered) train and test data, respectively. All equations using $Y_{t+1}$ apply analogously for $Y$ in non-sequential settings.
\subsection{Generative Model}
We use Conditional Variational Autoencoder (CVAE) as a generative predictor (GP) with an encoder and decoder components.
%where, the next location (Y) is conditioned by the sequence of past locations (X) and other features like the user id, and day of the week etc. 
%This comprises multilayer LSTM with self-attention encoder-decoder.
We use sparse-categorical cross-entropy loss ($\mathcal{L}_{\text{rec}}$) (depicted in Equation (\ref{eq:sp})) which represents the negative log probability of the Y, as true label in the training data given the input features (X), averaged over all samples (N) for sequence prediction tasks.

\( y_i \) represents the ith true level of the \( x_i \in X \)  sequence up to time step \( t-1 \):
$ y_i = f(X_{1:t-1})$ ; (for non-sequential data, $y_i$ simply corresponds to the label for $\mathbf{x}_i$).
\begin{equation}
\mathcal{L}_{\text{rec}}  = -\frac{1}{N} \sum_{i=1}^{N} \log p_{\theta}(y_i \mid \mathbf{x}_i) ; y_i\epsilon Y; x_i\epsilon X
\label{eq:sp}
\end{equation}
%We find this generative model outperforms the LSTM based discriminative predictor while predicting the next location.
 For prediction of binary-valued targets, we use binary cross-entropy loss $\mathcal{L}_{\text{bce}}$ (depicted in Equation (\ref{eq:bce}))
%    \begin{equation}
%    \mathcal{L}_{\text{bce}} = -\frac{1}{N} \sum_{i=1}^{N} \left[y_i \log p_{\theta}(y_i \mid \mathbf{x}_i) + (1 - y_i) \log (1 - p_{\theta}(y_i \mid \mathbf{x}_i)) \right]
%    \label{eq:bce}
%    \end{equation}
{\small
\begin{equation}
    \mathcal{L}_{\text{bce}} = -\frac{1}{N} \sum_{i=1}^{N} \left[
        y_i \log p_{\theta}(y_i \mid \mathbf{x}_i) + 
        (1 - y_i) \log (1 - p_{\theta}(y_i \mid \mathbf{x}_i))
    \right]
    \label{eq:bce}
\end{equation}
}
\vspace{-2mm}
   \begin{equation}
c_{\text{max}} = \max(Y_{\text{train}}) + 1
\label{eq:1}
\end{equation}
\textbf{Encoder:} 
%The encoder component comprises two LSTM layers. The first LSTM layer has 60 units in hidden states and returns the full sequence of output. Dropout is applied to this output to prevent overfitting. The output from this layer then goes to a self-attention layer with sigmoid activation, followed by the next LSTM layer with 40 units in hidden states, and finally goes to a latent space representation layer conditioned on X. 
The encoder learns the latent space representation (Z).
Z is obtained by computing the mean $\mu $ and log-variance $\sigma$ with the reparameterization trick to sample from the latent space, first $\epsilon$ is sampled from $N(0,1)$ and then z is computed as $z = \mu(Y|X) + \sigma ^{ 1/2}(Y|X)*\epsilon.$

\textbf{Decoder:} The decoder takes Z as input and repeats across the time step which is the maximum sequence length in the training data and combines this with the conditional input X, considering X as temporal data. Now this Z conditioned on X is passed to the next neural layers as used in Encoder. %LSTM layer with 40 units in hidden states, returns the full sequence of output. Dropout is applied to this output to prevent overfitting. The output from this layer then goes to a self-attention layer with sigmoid activation, followed by the next LSTM layer with 60 units in hidden states, and finally output of this LSTM goes to a dense layer. 
The final output of the decoder goes to a dense layer. This final dense layer has $c_{max}$ (defined in Equation~\eqref{eq:1}) number of nodes with softmax activation to predict the next location.
 
\textbf{Generator:} The decoder model is employed to generate new data samples of Y by passing the latent samples sampled from a Gaussian distribution and using conditional inputs.
For \textbf{sequence prediction tasks} such as next-location modeling, we use the following CVAE loss (Equation \ref{eq:vae_loss}). The reconstruction loss uses sparse categorical cross-entropy (Equation \ref{eq:sp}. We apply \textbf{KL annealing} \cite{klAnn}, kl\_weight (a gradually increasing weight) to multiply the KL divergence term to counter KL-vanishing during the initial training. 
\begin{equation}
\begin{split}
\mathcal{L}(\theta, \phi; \mathbf{Y}_t, \mathbf{X}_{t-1:t-n}) = 
\\
- \mathbb{E}_{q_\phi(\mathbf{z}|\mathbf{Y}_t, \mathbf{X}_{t-1:t-n})} 
\left[\log p_\theta(\mathbf{Y}_t|\mathbf{z}, \mathbf{X}_{t-1:t-n})\right] \\
+ \text{kl\_weight} \cdot \text{KL}\left(q_\phi(\mathbf{z}|\mathbf{Y}_t, \mathbf{X}_{t-1:t-n}) 
\parallel p_\theta(\mathbf{z}|\mathbf{X}_{t-1:t-n})\right)
\end{split}
\label{eq:vae_loss}
\end{equation}
$\theta$, $\phi$ are the parameters of the decoder, and  encoder network respectively. $\mathbf{Y}_t$ is the target output %( next locations),
$\mathbf{X}_{t-1:t-n}$ is the input sequence comprising different features.
   $q_\phi(\mathbf{z}|\mathbf{Y}_t, \mathbf{X}_{t-1:t-n})$ is the approximate posterior distribution,
    $p_\theta(\mathbf{Y}_t|\mathbf{z}, \mathbf{X}_{t-1:t-n})$ is the likelihood of the data given the latent variable and the conditional input.
    $p_\theta(\mathbf{z}|\mathbf{X}_{t-1:t-n})$ is the prior distribution of the latent variable given the conditional input.
    $\text{KL}$ denotes the Kullback-Leibler divergence \cite{KlD}.
%    This loss in Equation ~\eqref{eq:vae_loss} applies to sequence-based prediction tasks (e.g., next-location), where $\mathbf{X}_{t-1:t-n}$ denotes a sequence of features and $\mathbf{Y}_t$ the next item in the sequence.
For non-sequential prediction tasks the same loss Eq.~\eqref{eq:vae_loss} is used, where the reconstruction term follows binary cross-entropy (Eq.~\eqref{eq:bce}), with $y_i$ corresponding directly to the label for $\mathbf{x}_i$.
    
% To accommodate both sequence-based and non-sequential tasks, we define the loss in a general form where 
%X represents the conditioning input and 
%Y the prediction target. This formulation applies to both spatiotemporal sequence prediction (e.g., next-location) and binary classification (e.g., Asia dataset) without loss of generality.   
%For binary prediction task we use the following CVE loss.     
%\begin{equation}
%\begin{split}
%\mathcal{L}(\theta, \phi; \mathbf{Y}, \mathbf{X}) = 
%- \mathbb{E}_{q_\phi(\mathbf{z}|\mathbf{Y}, \mathbf{X})} 
%\left[ y \log p_\theta(y \mid \mathbf{z}, \mathbf{X}) + (1 - y) \log (1 - p_\theta(y \mid \mathbf{z}, \mathbf{X})) \right] \\
%+ \text{kl\_weight} \cdot \text{KL} \left( q_\phi(\mathbf{z}|\mathbf{Y}, \mathbf{X}) \parallel p_\theta(\mathbf{z}|\mathbf{X}) \right)
%\end{split}
%\label{eq:binary_cvae_loss}
%\end{equation}   
\subsection{Causal Sensitivity Identification}
\label{subsec:cs}
The proposed method determines the causal sensitivity considering two perspectives. 
%Throughout this section, all equations using $Y_{t+1}$ apply analogously for $Y$ in non-sequential settings.
\begin{enumerate}
\item \textbf{Causally sensitive feature ($Z_{CF}$) and reduction of confounding bias}: We propose the following steps to determine the causal sensitivity of a feature without considering any prior knowledge of causal graph. We aim to reduce the confounding effect by blocking the backdoor path (depicted in Figure \ref{bd}) which connects X and Y with at least one common ancestor or common cause \( X \leftarrow Z \rightarrow Y \), where X is not the cause of Z. %For example, regarding the particular day of the week (W) or the start time of the location visits(Smin).
%   We propose following method to identify causal sensitivity of a feature and to identify the cause effect relationship.
\begin{enumerate}
\item \textbf{Factual training}: Train the generative predictor (GP) model using the factual input \( X_{\text{train}}\) to create GP-F (Baseline) as shown in Equation \ref{eq:4}.% conditioning on \( F_{\text{CS}} \). 
\item \textbf{Intervention}: Apply the intervention, $( do(X=X_\text{altered}))$. Train the GP using \( X_{\text{train\_Altered}} \) to obtain GP-I. 
% Example: the highest frequent location ID is replaced with location ID 0, where \( X \supseteq LS_{\text{past\_visits}} \), conditioning on \( F_{\text{CS}} \) 
      \begin{equation}
    %P(Y_{t+1} \mid \text{do}(X_{t})) = \sum_{W} P(Y_{t+1} \mid X_{t}, F_{\text{CS}}) P(F_{\text{CS}})
    P(Y_{t+1} \mid (X_{t})) = \sum_{F_{\text{CS}}} P(Y_{t+1} \mid X_{t}) 
    \label{eq:4}
\end{equation}
\begin{figure}[!h]
    \centering
    \includegraphics[width=0.70\linewidth]{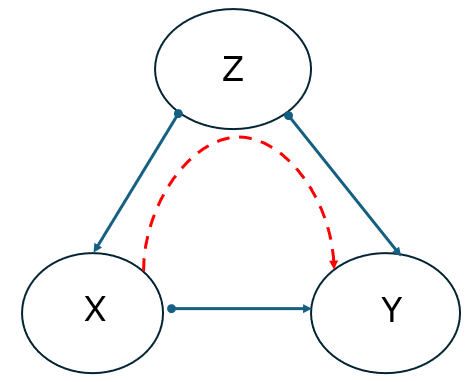}
    \caption{Causal graph depicting backdoor path}
    \label{bd}
\end{figure}
\begin{equation}
\scalebox{0.75}{$
    P(Y_{t+1} \mid \text{do}(X=X_{\text{altered}})) = 
    \sum_{F_{\text{CS}}} 
    P(Y_{t+1} \mid X_{\text{altered}}, F_{\text{CS}}) P(F_{\text{CS}})
$}
\label{eq:I}
\end{equation}
\item \textbf{Factual prediction}: Test the GP-F using \( X_{\text{test}} \).
\begin{equation} 
Y_{(t+1)\text{factual}} = \text{GP-F.decoder}(Z_{FC}, X_{\text{test}})
\label{eq:pf}
\end{equation}
\item \textbf{Interventional prediction}: Test the GP-I using \( X_{\text{test}} \).
\begin{equation}
Y_{(t+1)\text{Interventional}} = \text{GP-I.decoder}(Z_{IF}, X_{\text{test}})
\label{eq:pci}
\end{equation}
%\( X_{\text{test}} \) denotes the factual (unaltered) test data. All equations using $Y_{t+1}$ apply analogously for $Y$ in non-sequential settings.
 \item \textbf{Features having causal influence:} 
If the prediction error is higher (i.e., accuracy is lower) in the factual scenario (c) than in the interventional scenario (d), this indicates, that the feature is causally sensitive and acts as a common influencer, as blocking of backdoor path enables the direct causal path between X and Y and reduces the confounding bias in the causal effect of X on Y. We present this in terms of the difference in prediction accuracy, \(\Delta \text{Acc}\). (Examples \text{Acc}: \( \text{Acc}@1, \text{MRR} \in \text{Acc} \) defined in the section ~\ref{subsec:per} Performance Measure). 
 The difference in accuracy is defined as:
  \begin{equation}
    \Delta \text{Acc} = \text{Acc}_{\text{interventional}} - \text{Acc}_{\text{factual}} ;  \Delta \text{Acc} >0
    \label{eq:acc}
  \end{equation}
\end{enumerate}
\item \textbf{Counterfactuals to measure the change in effect when the cause has changed}
\begin{enumerate}
   \item \textbf{Obtain factual latent representation, and GP-F}: Get \( \mathbf{z_f} \): Train the GP using factual \( X_{\text{train}} \),  \( Y_{\text{train}} \).
\begin{equation}
    \mathbf{z_f} \sim q_\phi( {Y}_t , \mathbf{X}_{t-1:t-n})\_factual.
  \end{equation}
The trained encoder of GP-F is used to obtain \( Z_{\text{FC}} \), the factual latent representation from $X_{\text{test}}$
\[
Z_{\text{FC}} = \text{GP-F.encoder}(X_{\text{test}})
\] 
%Where, $Z_{\text{FC}}$ is the factual latent representation.
 \item  \textbf{Obtain counterfactual latent representation  \( Z_{\text{CF}} \):} \( Z_{\text{CF}} \) follows counterfactual probability which is computed based on the equation ~\eqref{eq:cf}
%\begin{equation}
% P(Y_{X=x'} \mid X=x, Y=y) \approx  \int P(Y \mid X=x', \mathbf{z}) \, q_\phi(\mathbf{z} \mid X=x, Y=y) \, d\mathbf{z}.
%\end{equation}  
\begin{equation}
\resizebox{\linewidth}{!}{$
    P(Y_{X = x'} \mid X = x, Y = y) \approx 
    \int P(Y \mid X = x', \mathbf{z}) \, q_\phi(\mathbf{z} \mid X = x, Y = y) \, d\mathbf{z}
$}
\label{eq:cf}
\end{equation}
   Obtain the counterfactual latent representation from the $X_{test\_altered}$.
    \begin{equation}
Z_{\text{CF}} = \text{GP-F.encoder}(X_{\textbf{test\_altered}})
    \label{eq:cf_latent}
\end{equation}
 \( Z_{\text{CF}} \) is the counterfactual latent representation.
    \item \textbf{Generate factual and counterfactual predictions:}
    Use the decoder of GP-F to predict the outcome \( Y_{\text{t+1}} \) for the factual scenario:
    \begin{equation} 
 Y_{(t+1)\text{factual}} = \text{GP-F.decoder} (Z_{FC} , X_{\text{test}})
  \label{eq:pf}
    \end{equation}
    %where, \( Z_{FC} \) is the latent representation for the factual scenario.
   \begin{equation} 
    Y_{(t+1)\text{counterfactual}} =  \text{GP-F.decoder}(Z_{CF} , X_{\textbf{test\_altered}})
    \label{eq:pcf}
    \end{equation}   
\end{enumerate} % for counterfactual
\end{enumerate} 
We use counterfactuals as described above on the test data and also generate counterfactuals $Y_{(t+1)\text{counterfactual}}$. The difference in accuracy \hypertarget{cfac}{between counterfactual and factual scenarios \( \Delta \text{Acc} < 0} (\Delta \text{Acc} = \text{Acc}_{\text{counterfactual}} - \text{Acc}_{\text{factual}}   
\)), signifies causal path \( X \rightarrow Y \). The proposed generative causal sensitivity identification method is presented in Algorithm \ref{alg:causal_sensitivity}, without using any prior knowledge of causal graph and applying any causality constraints during learning.
\begin{algorithm}[!h]
\caption{Causal Sensitivity Identification Method}
\label{alg:causal_sensitivity}
\textbf{Input}: $X_{\text{train}}, Y_{\text{train}}, X_{\text{train\_altered}}, X_{\text{test}}, X_{\text{test\_altered}}$\\
\textbf{Output}: $\Delta \text{Acc}$ , Causally sensitive features, Causal path, Counterfactual prediction
\begin{algorithmic}[1]
\State Train GP-F on $X_{\text{train}}, Y_{\text{train}}$ to learn encoder $q_\phi$ and decoder $p_\theta$ $\rightarrow$ Eq.~\eqref{eq:vae_loss}
\State Train GP-I on $X_{\text{train\_altered}}$ $\rightarrow$ Eq.~\eqref{eq:I}
\State Compute factual latent: \\
          $Z_{\text{FC}} = \text{GP-F.encoder}(X_{\text{test}})$
\State Compute interventional latent: \\
         $Z_{\text{IF}} = \text{GP-I.encoder}(X_{\text{test}})$
\State Factual prediction: \\
         $Y_{(t+1)\text{factual}} = \text{GP-F.decoder}(Z_{\text{FC}}, X_{\text{test}})$ $\rightarrow$ Eq.~\eqref{eq:pf}
\State Interventional prediction: \\
        $Y_{(t+1)\text{interventional}} = \text{GP-I.decoder}(Z_{\text{IF}}, X_{\text{test}})$ $\rightarrow$ Eq.~\eqref{eq:pci}
\State $\Delta \text{Acc} = \text{Acc}_{\text{interventional}} - \text{Acc}_{\text{factual}}$ $\rightarrow$ Eq.~\eqref{eq:acc}
\If{$\Delta \text{Acc} > 0$}
    \State Feature is causally sensitive
\EndIf
\State Counterfactual latent: $Z_{\text{CF}} = \text{GP-F.encoder}(X_{\text{test\_altered}})$ $\rightarrow$ Eq.~\eqref{eq:cf_latent}
\State Counterfactual prediction: $Y_{(t+1)\text{counterfactual}} = \text{GP-F.decoder}(Z_{\text{CF}}, X_{\text{test\_altered}})$ $\rightarrow$ Eq.~\eqref{eq:pcf}
\State $\Delta \text{Acc} = \text{Acc}_{\text{counterfactual}} - \text{Acc}_{\text{factual}}$ $\rightarrow$ Eq.~\eqref{eq:cf}
\If{$\Delta \text{Acc} < 0$}
    \State Infer causal path: $X \rightarrow Y$
\EndIf
\end{algorithmic}
\end{algorithm}
\subsection{Causally sensitive recommendation/prediction:}
%The proposed generative causal sensitivity identification method presented in Algorithm \ref{alg:causal_sensitivity} is capable of identifying causally sensitive features, $F_{CS}$, and assessing the impact of changes in the cause on its effect and finding the causal path using  a CVAE based generative model. Notably this is achieved without using any prior knowledge of causal graph and applying any causality constraints during learning. 
The identified $F_{CS}$ (Algorithm \ref{alg:causal_sensitivity}) is applied to condition the prediction task.% following equation Eq.~\eqref{eq:pf}, ($Y_{(t+1)\text{factual}} = \text{GP-F.decoder}(Z_{\text{FC}}, X_{\text{test}})$). 
We refer to this as generative causally sensitive prediction (\textbf{GCSP}) presented in Algorithm \ref{alg:gcsp}.
\begin{algorithm}[!h]
\caption{Generative Causally Sensitive Prediction (GCSP)}
\label{alg:gcsp}
\textbf{Input:} Factual training data $(X_{\text{train}}, Y_{\text{train}})$, test data $X_{\text{test}}$ \\
\textbf{Output:} Prediction $Y_{(t+1)\text{factual}}$ using causally sensitive conditioning
\begin{algorithmic}[1]
\State Train GP-F (Generative Predictor - Factual) using CVAE on $(X_{\text{train}}, Y_{\text{train}})$
\State Identify causally sensitive features $F_{CS}$ via intervention analysis (Algorithm~\ref{alg:causal_sensitivity})
\State Condition the model on $F_{CS}$ %to reduce confounding bias
\State Encode test data to obtain factual latent representation:
    \[
    Z_{\text{FC}} = \text{GP-F.encoder}(X_{\text{test}})
    \]
\State Generate predictions using decoder conditioned on $F_{CS}$:
    \[
    Y_{(t+1)\text{factual}} = \text{GP-F.decoder}(Z_{\text{FC}}, X_{\text{test}}) following Eq.~\eqref{eq:pf}
    \]
\State \Return $Y_{(t+1)\text{factual}}$ as the causally conditioned next prediction
\end{algorithmic}
\end{algorithm}

Our method integrates causal sensitivity identification with generative prediction and is applicable to both general and sequential prediction tasks, such as next-location prediction. In contrast to prior works, it addresses causal impact analysis with the following key features:
%\begin{itemize}
%    \item Identification of causally sensitive features through interventional analysis, and quantification of their influence using a generative predictor;
%    \item  Assessing the impact of changes in causes on the predicted outcome, and identifies the causal path
%    \item A unified prediction framework that operates without prior knowledge of the causal graph or structural constraintsts (such as acyclicity, as required in methods like NOTEARS~\cite{NOTEARS}).
%\end{itemize}
%Our method integrates causal sensitivity identification with generative prediction and is applicable to both general and sequential prediction tasks, such as next-location prediction. In contrast to prior works, our approach addresses causal impact analysis with the following key features:
\begin{itemize}
    \item Identification of causally sensitive features through interventional analysis and quantification of their influence using a generative predictor;
    \item Assessment of the impact of changes in causes on predicted outcome, enabling identification of causal path;
    \item A unified prediction framework that operates without prior knowledge of the causal graph or structural constraints (such as acyclicity, as required in methods like NOTEARS~\cite{NOTEARS}).
\end{itemize}
\section{Evaluation of Proposed Method}
In this section, we demonstrate causal sensitivity identification using the proposed method on the \textbf{Asia} dataset~\cite{lauritzen1988local} and on the \textbf{GeoLife} \cite{geolife} data.
%\paragraph{Scope of Causal Path Evaluation.}
%It is important to note that for the Asia dataset, the underlying causal structure is known, which enables explicit validation of the causal paths identified by our method. In contrast, for the GeoLife dataset, the ground-truth causal graph is not available. Therefore, our analysis for GeoLife focuses on the identification of causally sensitive features and the evaluation of their impact on prediction performance, rather than on validating specific causal paths.

It is important to note that the underlying causal structure for the Asia dataset is known, which allows for explicit validation of the causal paths identified by our method. In contrast, the GeoLife dataset does not have a ground-truth causal graph available for validation.
\subsection{Causal Sensitivity Identification on The Asia Dataset}
\label{subsec:asia}
To demonstrate the proposed causal sensitivity identification method we first apply it to the \textbf{Asia} dataset. 

\textbf{Data}: Asia real world data set of Bayesian Network Repository (BnLearn) \cite{bnlearn} contains 8 binary variables (e.g., \texttt{smoke}, \texttt{lung}, \texttt{bronc}, \texttt{dysp}). The directed acyclic graph (DAG) structure of the Asia data is predefined and presents known causal relationships among the variables, and is widely used in causal learning and inference tasks.

\textbf{Causal sensitivity analysis:}
We focus our causal sensitivity analysis on predicting the target variable \texttt{dysp} (shortness of breath) while evaluating how conditioning on subsets of features and intervening on variables like \texttt{either} affects predictive accuracy. Our method is used to identify minimal sufficient sets, isolate confounders such as \texttt{smoke}, and study performance shifts under interventions like \texttt{do(either=1)}.
We implement a CVAE to model the distribution of the binary target variable \texttt{dysp} conditioned on various subsets of input variables. Generative predictor CVAE is trained using binary cross-entropy loss combined Eq.~\eqref{eq:bce} with a KL divergence regularization term. The encoder takes as input the conditioning features (e.g., \texttt{either}, \texttt{smoke}, \texttt{bronc}) and the target \texttt{dysp}, and outputs the latent mean and log-variance. The decoder reconstructs \textbf{dysp} from samples drawn from the latent space and the same conditioning input. 
 
We use a multi-layer perceptron (MLP) based encoder with 16 hidden units to map the concatenated input of the target variable and conditioning variables ($[target, conditioning features]$) map to a latent space of dimension 2. The model is trained for 400 epochs using the Adam optimizer with a learning rate of $10^{-3}$, optimizing the binary cross-entropy reconstruction loss along with a KL divergence regularization term.
%By intervening on specific variables such as \texttt{either} (i.e., applying \texttt{do(either=1)}), we evaluate the sensitivity of the model's prediction and identify the minimal sufficient conditioning set required to maximize prediction accuracy.
\begin{figure}[!ht]
    \centering
    \includegraphics[width=0.60\columnwidth]{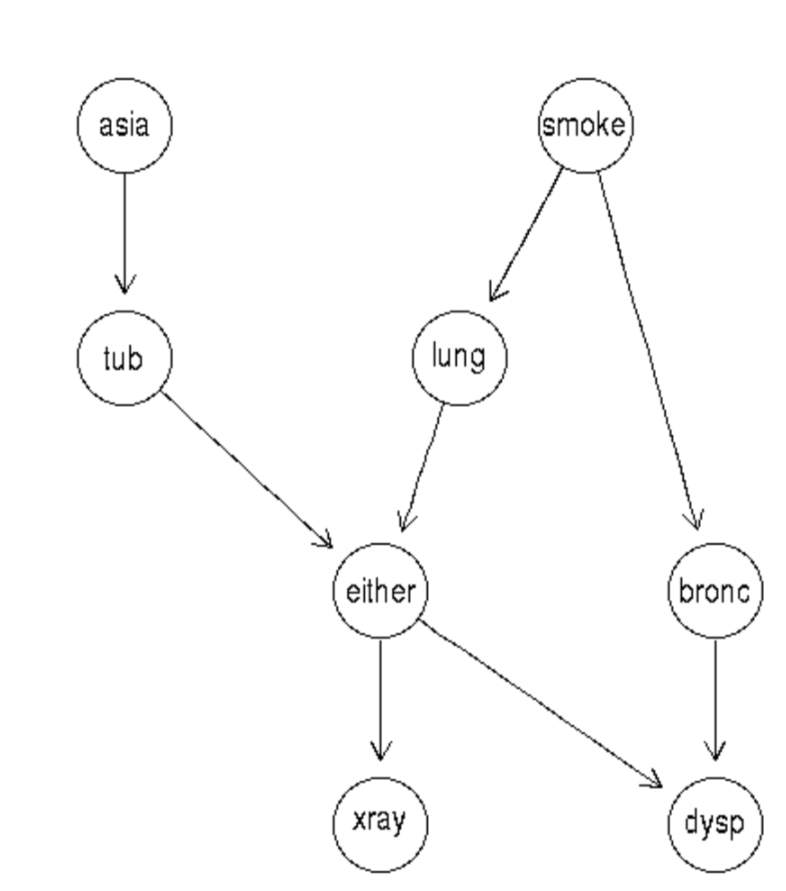}
    \caption{Causal graph - Asia \protect \cite{ke2020learning}}
    \label{fig:asia}
\end{figure}

\begin{table}[!h]
\centering
\scriptsize
\caption{Asia: Identification of causally sensitive features for \texttt{dysp} under factual and interventional (Intv) scenarios using the proposed method.}
\begin{tabular}{lccc}
\hline
\textbf{Conditioning Set} & \textbf{Scenario} & \textbf{Accuracy} \\
\hline
\texttt{[either]} & Factual & 0.660 \\
\texttt{[either]} & Intv & 0.580 \\
\texttt{[either, bronc]} & Factual & 0.815 \\
\texttt{[either, bronc]} & Intv & 0.850 \\
\texttt{[either, bronc, lung]} & Factual & 0.825 \\
\texttt{[either, bronc, lung]} & Intv & 0.845 \\
\texttt{[either, bronc, lung, tub]} & Factual & 0.835 \\
\texttt{[either, bronc, lung, tub]} & Intv & 0.845 \\
\texttt{[either, smoke, bronc]} & Factual & 0.850 \\
\texttt{[either, smoke, bronc]} & Intv & \textbf{0.855} \\
\texttt{[either, smoke, bronc, tub]} & Factual & 0.850 \\
\texttt{[either, smoke, bronc, tub]} & Intv & 0.850 \\
\texttt{[either, smoke, bronc, lung]} & Factual & 0.845 \\
\texttt{[either, smoke, bronc, lung]} & Intv & 0.850 \\
\texttt{[either, smoke, bronc, lung, tub]} & Factual & 0.850 \\
\texttt{[either, smoke, bronc, lung, tub]} & Intv & 0.850 \\
\hline\end{tabular}
\label{tab:dysp_accuracy_final}
\end{table}
%\begin{table}[ht]
%\centering
%\caption{Counterfactual Sensitivity for Target \texttt{dysp} under Interventions}
%\begin{tabular}{lccc}
%\toprule
%\textbf{Intervened Variable} & \textbf{Factual Accuracy} & \textbf{Counterfactual Accuracy} & \textbf{Delta} \\
%\midrule
%either & 0.83 & 0.59 & \textbf{-0.240} \\
%smoke  & 0.83 & 0.84 & +0.010 \\
%bronc  & 0.83 & 0.365 & \textbf{-0.465} \\
%lung   & 0.83 & 0.85 & +0.020 \\
%tub    & 0.83 & 0.855 & +0.025 \\
%\bottomrule
%\end{tabular}
%\label{tab:dysp_sensitivity}
%\end{table}
%
\begin{table}[!h]
\centering
\caption{Asia: Counterfactual (CF) sensitivity analysis for \texttt{dysp} using the proposed method.}
\begin{tabular}{lccc}
%\begin{tabular}{ccc}
\hline
\textbf{Counterfactual} & \textbf{Factual Acc.} & \textbf{CF Acc.} & \textbf{Delta} \\
\hline
either & 0.83  & 0.59  & \textbf{--0.240} \\
smoke  & 0.83  & 0.84  & +0.010 \\
bronc  & 0.83  & 0.365 & \textbf{--0.465} \\
lung   & 0.83  & 0.85  & +0.020 \\
tub    & 0.83  & 0.855 & +0.025 \\
\hline
\end{tabular}
\label{tab:dysp_cf}
\end{table}
Based on the proposed method (Algorithm~\ref{alg:causal_sensitivity}), we evaluate the causal sensitivity of the features on the prediction of the target variable \texttt{dysp} using the Asia dataset presented in Table \ref{tab:dysp_accuracy_final}. Our results indicate that \texttt{bronc} and \texttt{smoke} are causally sensitive variables for \texttt{dysp}, as there is a significant improvement in the accuracy of interventional scenarios. Furthermore, conditioning \texttt{smoke} with \texttt{bronc} while intervening do(either = 1) gives the highest performance indicates the possibility of \texttt{smoke} being a confounder, contributing to backdoor paths like \texttt{smoke}~$\rightarrow$~\texttt{bronc}~$\rightarrow$~\texttt{dysp} and \texttt{smoke}~$\rightarrow$~\texttt{lung}~$\rightarrow$~\texttt{either}~$\rightarrow$~\texttt{dysp}. The ground truth causal graph is depicted in figure \ref{fig:asia}.
 
In Table \ref{tab:dysp_cf} we present our findings highlighting the utility of counterfactual inference in uncovering both direct and indirect causal relationships without prior knowledge of the underlying graph. Specifically, intervening on \texttt{bronc} and \texttt{either} resulted in a decline in prediction accuracy by $-0.465$ and $-0.240$, respectively, thereby validating the existence of direct causal paths: \texttt{bronc}~$\rightarrow$~\texttt{dysp} and \texttt{either}~$\rightarrow$~\texttt{dysp}.
%during the learning phase. % and performance gains in dynamic, user-centric environments.
We compare our method with prior approaches such as Ke et al.~\cite{ke2020learning}, who apply a neural causal model to the Asia dataset and successfully identify key paths. %such as \texttt{smoke} $\rightarrow$ \texttt{bronc} $\rightarrow$ \texttt{dysp} and \texttt{smoke} $\rightarrow$ \texttt{lung} $\rightarrow$ \texttt{either} $\rightarrow$ \texttt{dysp}.
Their method achieves high structural accuracy without requiring knowledge of intervention targets. However, their method focuses primarily on recovering the causal graph structure and does not explicitly quantify the effect of individual variables on prediction outcomes.

In contrast our generative approach identifies causally sensitive features and their effects on the prediction of \texttt{dysp} through performance deviations under factual, interventional, and counterfactual scenarios, and identifies the direct causal path, validating its effectiveness in capturing structural and functional causal dependencies.
%In contrast, our generative approach not only uncovers causally sensitive features but also identifies direct causal paths by quantifying their impact on prediction accuracy under counterfactual settings
%Our focus on task-specific causal sensitivity via interventions and counterfactuals.
%
%Use generative models (CVAE) to assess predictive performance under causal manipulations.
%
%Can handle both structured data (Asia) and sequential, unstructured data (GeoLife).

We further compare our method against CausalVAE \cite{Yang2021}.
\begin{table}[!h]
\centering
\caption{Asia: counterfactual sensitivity analysis for \texttt{dysp} applying CausalVAE. }
\begin{tabular}{lccc}
\hline
\textbf{Counterfactual} & \textbf{Factual Acc.} & \textbf{CF Acc.} & \textbf{Delta} \\
\hline
%asia   & 0.62 & 0.62  & 0.000  \\
tub    & 0.62 & 0.62  & 0.000  \\
smoke  & 0.62 & 0.62  & 0.000  \\
lung   & 0.62 & 0.62  & 0.000  \\
bronc  & 0.62 & 0.62  & 0.000  \\
either & 0.62 & 0.62  & 0.000  \\
%xray   & 0.62 & 0.62  & 0.000  \\
\hline
\end{tabular}
\label{tab:causalvae_cf}
\end{table}

Table~\ref{tab:causalvae_cf} presents the counterfactual evaluation of CausalVAE on the Asia dataset for predicting \texttt{dysp}. Notably, the model shows no significant variation in accuracy across counterfactual scenarios, indicating a lack of sensitivity to causal structure. This contrasts with our method (Table~\ref{tab:dysp_cf}), which identifies \texttt{bronc} and \texttt{either} as causally sensitive features. This supports the claim that our approach better captures the underlying causal relationships necessary for meaningful counterfactual reasoning.
%\begin{table}[t]
%\centering
%\caption{Causal Discovery Performance on the \texttt{Asia} Dataset}
%\begin{tabular}{|l|c|c|c|}
%\hline
%\textbf{Method} & \textbf{SHD} & \textbf{TPR} & \textbf{FDR} \\
%\hline
%GES & 3 & 0.875 & 0.22 \\
%Enhanced GES & 4 & 1.000 & 0.33 \\
%NOTEARS & 4 & 0.500 & 0.33 \\
%Enhanced NOTEARS & 5 & 0.375 & 0.25 \\
%KCRL & 4 & 0.500 & 0.20 \\
%Enhanced KCRL & 2 & 0.750 & 0.00 \\
%\textbf{CVAE-Causal (Ours)} & \textbf{3} & \textbf{0.667} & \textbf{0.769} \\
%\hline
%\end{tabular}
%\label{tab:asia_causal_comparison}
%\end{table}

%We compare our method against prior work that explicitly targets causal path identification under unknown interventions. Notably, Ke et al.~\cite{ke2020learning} evaluate their neural causal model on the Asia dataset and demonstrate the ability to recover key causal paths such as \texttt{smoke} $\rightarrow$ \texttt{bronc} $\rightarrow$ \texttt{dysp} and \texttt{smoke} $\rightarrow$ \texttt{lung} $\rightarrow$ \texttt{either} $\rightarrow$ \texttt{dysp}. Their method achieves high structural accuracy without requiring explicit knowledge of intervention targets. In contrast, our approach identifies causally sensitive features and assesses their impact on the prediction of \texttt{dysp} through performance deviations observed under factual, interventional, and counterfactual scenarios. This generative modeling framework offers a complementary perspective that captures both structural and functional causal dependencies, and empirically recovers all ground-truth causal paths in the Asia network.

% Include Table 1 and Table 2 here.
\subsection{Causal Sensitivity Identification on The GeoLife Data}
We apply the proposed method to predict the next location of human trajectory using GeoLife \cite{geolife} data. 
%The prediction of the next location is constructed as a variable length sequence prediction problem considering the trajectories, time series of track-points $<la,lo,t>$ of individuals, where ($la,lo$) is the spatial GPS coordinates like latitude, longitude, and t is the time stamp with the classification objective. 
%\textbf{Data:}

\textbf{Data:} GeoLife is a human trajectory dataset collected by 182 users in a period of over three years (from April 2007 to August 2012) under the GeoLife project, Microsoft Research Asia. This comprises the GPS trajectory a sequence of time-stamped points, each of which contains the information of latitude, longitude and altitude having diverse sampling rate. This dataset has 17,621 trajectories covering a total distance of 1.2 million kilometers and more than 48,000 hours of duration. This trajectory dataset includes a wide range of users with diverse outdoor movements, like shopping, sightseeing, dining etc., along with their life routines like go home and go to work.

\textbf{Factual data:} The unaltered GeoLife data is exploited for factual analysis. 

\textbf{LS:}  The sequence of past location visits.

 \textbf{Altered data:}  We create altered versions of the data by modifying the sequence of location visits to conduct intervention and counterfactual analysis, as follows:
  
 \textbf{$LS_1$:} Replace the most frequently visited location ID with the third most frequent.
  
\textbf{$LS_1$:}  Replace the most frequently visited location ID with location ID 0.

We follow these steps to exploit the proposed method and perform experimental analysis: 
\begin{enumerate}
\item Preprocessing of data 
\item High level feature extraction
\item Evaluate causal sensitivity identification using interventions, counterfactuals
\item Next location prediction in GPS trajectory.
%\item Ablation Study
\end{enumerate}
\subsubsection{Preprocessing of Data}
%We formulate the next location prediction task as a sequence prediction problem considering the trajectories, time series of track-points $<la,lo,t>$  of individuals,  where ($la,lo$) is the spatial GPS coordinates like latitude, longitude, and t is the time stamp.
%In this step trajectories are processed to extract staypoints and locations. Staypoints are a subset of trajectories where the user stays for a minimum duration of time. We follow \cite{trackintel} for the preprocessing to form locations by clustering the staypoints applying DBSCAN (Density-Based Spatial Clustering of Applications with Noise) \cite{dbscan}, with parameters $\epsilon = 20 $ controls the distance of which nearby staypoints will be merged into a single location and $min\_samples = 2 $ determines the minimum number of staypoints to form a location (i.e., number of visits needed at the same place to consider it as significant). We consider the sequence of past location visits and the extracted high-level mobility features with the embedding \cite{ref_article1} like the duration of stay, day of the week and start time encompassing both temporal and spatial characteristics along with user ID. We use the preprocessed data to identify the causal-sensitivity and there after predicting the next location in the trajectory. 
In this step trajectories are processed to extract staypoints and locations. Staypoints are a subset of trajectories where the user stays for a minimum duration of time. We follow \cite{trackintel} for the preprocessing to form locations.  %by clustering the staypoints applying DBSCAN (Density-Based Spatial Clustering of Applications with Noise) \cite{dbscan}, with parameters $\epsilon = 20 $ controls the distance of which nearby staypoints will be merged into a single location and $min\_samples = 2 $ determines the minimum number of staypoints to form a location (i.e., number of visits needed at the same place to consider it as significant). We consider the sequence of past location visits and the extracted high-level mobility features with the embedding \cite{ref_article1} like the duration of stay, day of the week and start time encompassing both temporal and spatial characteristics along with user ID. 
We use the preprocessed data to identify the causal-sensitivity and there after predicting the next location in the trajectory. 
\subsubsection{High Level Feature Extraction}

We use the open source Python library Trackintel \cite{trackintel} to process and analyze the GeoLife movement data as considered by authors in \cite{ref_article1} and extract the various high-level mobility features.
The high-level features considered are as follows:

Unique user identifier(UID), sequence of past location visits(LS), activity duration(DS), start minute (Smin), day of the week(W). 
The Smin feature adds a finer level of temporal granularity by indicating the specific start times of activity (location visit) within an hour or day. Feature W adds the perspective of the user’s daily life visits and the other out-of-routine location visits. 

\subsubsection{Performance Measure}
\label{subsec:per}
The following performance metrics are used: 

\textbf{Accuracy (Acc@k):} Measures how often the true location is in the top-k predictions. 
\begin{itemize}
    \item \( P \in \mathbb{R}^{N \times C} \): Predicted probabilities matrix (\(N\): samples, \(C\): classes),
    \item \( y \in \{1, \ldots, C\}^N \): is the vector of true labels.
\end{itemize}
%Top-k Accuracy is:
\[\text{Top-k Accuracy} = \frac{1}{N} \sum_{i=1}^{N} \mathbb{1}\left(y_i \in \text{Top-k}(P_i)\right)
\]
where \( \text{Top-k}(P_i) \) is the set of \( k \) classes with the highest predicted probabilities for sample \( i \), and \( \mathbb{1}(\cdot) \) is 1 if true, 0 otherwise. We report \textbf{\( \text{Top-k Accuracy} \times 100\%\)}.

\textbf{Mean Reciprocal Rank (MRR):} Computes the average reciprocal rank of the true label:
\[\text{Reciprocal Rank}(i) = \frac{1}{\operatorname{rank}_i(y_i)}, \quad
\]
 \[
  \text{MRR} = \frac{1}{N} \sum_{i=1}^{N} \frac{1}{\operatorname{rank}_i(y_i)}
\]
where \( \operatorname{rank}_i(y_i) \) is the position of \( y_i \) in the sorted predicted probabilities (\( \text{rank}=1 \) for the highest probability). We report \textbf{\( \text{MRR} \times 100\%\)}.

\textbf{Jensen-Shannon Divergence(JSD):} Measures the similarity between two probability distributions P and Q.
\[
\text{JSD}(P \| Q) = \frac{1}{2} D_{\text{KL}}(P \| M) + \frac{1}{2} D_{\text{KL}}(Q \| M), 
\]  
\[\quad \text{where } M = \frac{1}{2}(P + Q)
\]
The terms \( D_{\text{KL}}(P \| M) \) and \( D_{\text{KL}}(Q \| M) \) represent the KL divergence of \( P \) and \( Q \) with respect to \( M \), respectively. The JSD is symmetric and bounded between 0 and 1, with lower values indicating higher similarity between \( P \) and \( Q \).
\begin{figure*}[!h]
\centering
% Left subfigure
\begin{subfigure}[b]{0.48\textwidth}
    \centering
    \includegraphics[width=\linewidth]{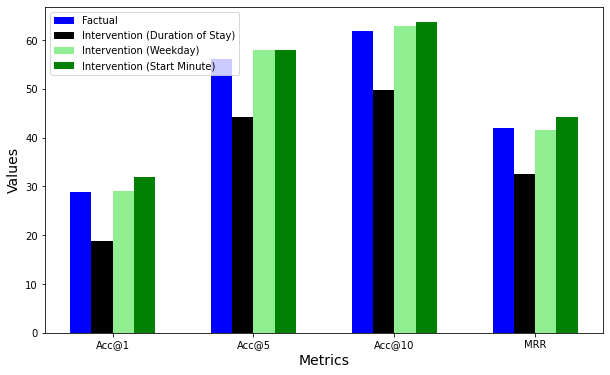}
    \caption{Causally sensitive features}
    \label{fig:fig2a}
\end{subfigure}
\hfill
% Right subfigure
\begin{subfigure}[b]{0.42\textwidth}
    \centering
    \includegraphics[width=\linewidth]{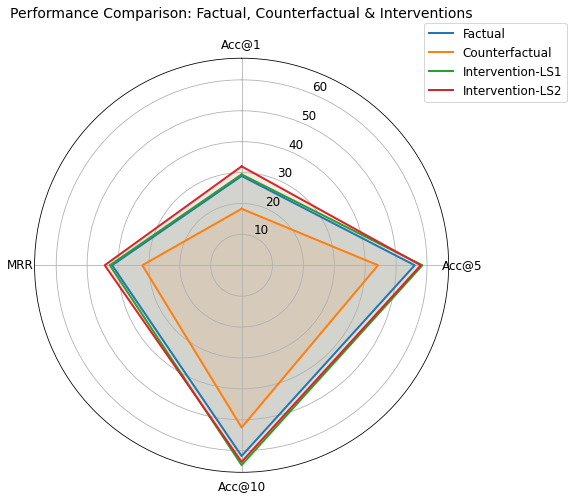}. %bar-plot3.png
    \caption{Performance: Factual (F), Intervention (I), Counterfactual (C)}
    \label{fig:count}
\end{subfigure}
\caption{Causal Sensitivity Identification: a: Causally sensitive features, b: Performance across factual, intervention, and counterfactual scenarios.}
\label{fig:causal_eval}
\end{figure*}
\begin{figure*}[!h]
\centering
% Left Subfigure
\begin{subfigure}[b]{0.42\textwidth}
    \centering
    \includegraphics[width=\linewidth]{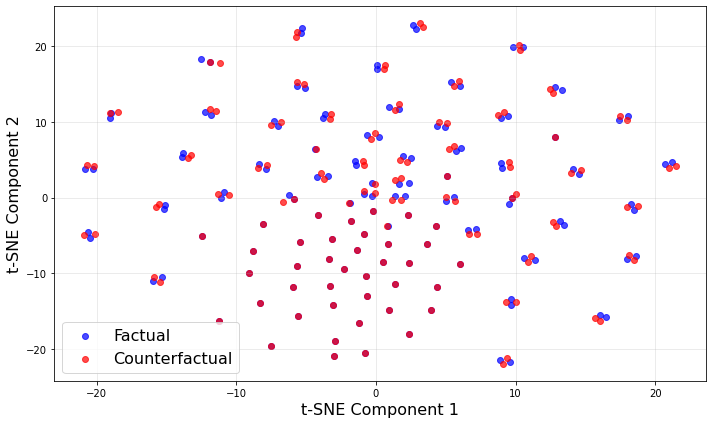}
    \caption{t-SNE visualization of factual ($Z_{FC}$) and counterfactual ($Z_{CF}$) latent spaces \cite{tsne}}
    \label{fig:dist_user21}
\end{subfigure}
\hfill
% Right Subfigure
\begin{subfigure}[b]{0.36\textwidth}
    \centering
    \includegraphics[width=\linewidth]{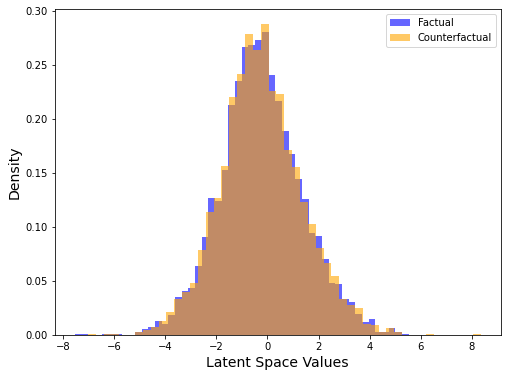}
    \caption{Distribution comparison using JSD between $Z_{FC}$ and $Z_{CF}$ (JSD = 0.4244)}
    \label{fig:dist_user9}
\end{subfigure}
\caption{Comparison of latent spaces $Z_{FC}$ and $Z_{CF}$ from the GP-F model: (a) shows the t-SNE projection; (b) shows the Jensen-Shannon divergence between their distributions.}
\label{fig:comparison_tsne_distribution}
\end{figure*}

\subsubsection{Results} 
We conduct extensive experimental analysis to validate our method. Results are obtained considering the data for the 45 selected users in GeoLife as considered by the authors in MHSA \cite{ref_article1}. These users have observation periods of more than 50 days to provide a longer observational time for getting a meaningful temporal pattern. Average accuracy Acc@k where k = 1, 5, 10 and average MRR values are computed across the users. JSD measures the similarity between factual and counterfactual latent space distribution.

\textbf{Generative model configuration:} Our method exploits CVAE with multilayer LSTM-based architecture and self-attention modules.

\textbf{Encoder:} The encoder comprises two LSTM layers. First one has 60 hidden units and outputs the full sequence, which is regularized by dropout layer, (dropout = 0.3) to prevent overfitting. The output from this layer is passed through a \textbf{self-attention} layer with sigmoid activation, followed by the second LSTM layer with 40 hidden units, the final output is mapped to a latent space (Z)  conditioned on X. 

\textbf{Decoder:} The decoder takes Z as input and repeats it across the maximum sequence length in the training data. This Z, conditioned on X is passed to the LSTM layer with 40 hidden units, which outputs the full sequence. Dropout layer (dropout = 0.3) is applied to the output. The output from the dropout layer is passed to a \textbf{self-attention} layer with sigmoid activation, followed by the next LSTM layer with 60 hidden units, and finally, the output from this LSTM is passed to a dense layer with $c_{max}$ (defined in Equation~\eqref{eq:1}) units of nodes with softmax activation to predict the next location. The reconstruction loss uses sparse categorical cross-entropy. 

We apply \textbf{KL annealing}, with kl\_weight (a gradually increasing weight) as described in the methodology we consider kl\_start epoch as 10, so up to 10 epochs the learning focuses only on the reconstruction error, kl\_annealtime = 20. We use Adam optimizer, batch size = 32, latent dimension 50, and train our model with 500 epochs.

%\hyperlink{cs}{causal features}
%\subsubsection{Results: Causal Sensitivity Identification:}
\begin{enumerate}
\item \textbf {Causally sensitive variable and establishing cause-effect relationship:}
\begin{itemize}
 \item We obtain GP-F, the factual baseline model following equation \eqref{eq:vae_loss} considering only LS, and measure the performance using factual test data. 
 \item We intervene LS in $X\_train$ by replacing the highest occurring location in LS with LS1 and LS2, resulting as \( X_{\text{train\_Altered}}\),  and then train the interventional model, GP-I, using \( X_{\text{train\_Altered}}\) conditioning on different candidate features (day of week  $W$, start time $Smin$, and duration of stay DS) to be identified as causally sensitive, denoted as $F_{\text{CS}}$.%as depicted in Figure \ref{fig:cg}. By following equation \eqref{eq:I} and conditioning on different features. We aim to mitigate the confounding bias by blocking the backdoor path, and isolate the causal effect of LS on Y, assuming $F_{\text{CS}}$ is a confounder.  
% \item We obtain GP-I, where we train the GP (Generative Predictor) using \( X_{\text{train\_Altered}}\) conditioning on different candidate features (day of week  $W$, start time $Smin$, and duration of stay DS). 
 \item We compute the performance of  both GP-F and GP-I on $X_{test}$ to assess changes in next-location prediction performance under intervention.
 % \item We compute the performance of next location prediction using $X_{test}$ using GP-I.
 \item We evaluate W, Smin, and DS as $F_{\text{CS}}$ considering equation \eqref{eq:acc}.
 \end{itemize}
In this scenario, as discussed in section ~\ref{subsec:cs}, conditioning on $Smin$ and $W$, we observe improvement in the best average performance, across all users, i.e.,  $\Delta \text{Acc} >0$, (equation \eqref{eq:acc}), establishing them as causally sensitive features representing the causal path as \( LS \leftarrow F_{\text{CS}} \rightarrow Y \), where \( F_{\text{CS}} \) acts as a common cause for \( LS \) and \( Y \).  

For duration of stay (DS), we observe average $\Delta \text{Acc}<0$; and for the best average performance approximately equal. 
This indicates DS does not have significant causal influence to the LS and the Y next location. Figure ~\ref{fig:fig2a} depicts the obtained results. %of causal sensitivity.

\item \textbf{Measure the change in effect when its cause has changed:}
\begin{itemize}
\item We use GP-F, the factual baseline model, to encode the factual test data $X_{\text{test}}$ and obtain the latent representation $Z_{FC}$.
\item To assess counterfactual effects, we alter the sequence of past location visits in $X_{\text{test}}$ to form $X_{\text{test\_Altered}}$ (Equation~\eqref{eq:cf}) and compute the counterfactual latent $Z_{\text{CF}}$ (Equation~\eqref{eq:cf_latent}).
\item Using these, we generate both factual (Equation~\eqref{eq:pf}) and counterfactual (Equation~\eqref{eq:pcf}) predictions, representing alternate trajectory outcomes.
\item We compare performance metrics of factual and counterfactual predictions to evaluate the impact of changes in causes.
\end{itemize}

Figure ~\ref{fig:count} depicts the counterfactual(C), factual(F) and interventional(I) scenarios where we find the average performance of counterfactual scenario is less than the Factual best one As discussed in section ~\ref{subsec:cs} i.e., $\Delta \text{Acc} < 0$ in this scenario.                                                                                                                                                                                                                                                                                                                                                                                                                         This evolution helps to measure the change in effects on the next location visit when the sequence of previous location visits has changed, this further helps to generate the possible alternate trajectories.
Figure~\ref{fig:comparison_tsne_distribution} depicts the divergence of factual and counterfactual latent space distribution. 

\item \textbf{Causally sensitive prediction of next location of the trajectory:}

%We evaluate the proposed CVAE-based generative predictor with a \textbf{multilayer-LSTM-self-attention} encoder and decoder using factual data and conditioning on the causally sensitive feature, Smin. We generate $n = 20$ samples and consider the best-performing prediction among them. %We train this model with 500 epochs and batch size = 32 with dropout rate = 0.3. To compute $kl\_weight$ as described in the methodology we consider kl\_start epoch as 10, so up to 10 epochs the learning focuses only on the reconstruction error, kl\_annealtime = 20.
We evaluate the proposed causally sensitive generative predictor using factual data and conditioning on the causally sensitive feature, Smin. For each instance, we generate $n = 20$ samples and report the best-performing prediction.
We summarize in Table \ref {tab:combined_Results} the obtained results using factual GeoLife data along with the results of relevant state-of-the-art (SoA) as discussed in the related work \hyperlink{trajcomp}{\texttt{Our emphasis}}. 

We compare the best-performing results from SoA methods trained on combined user data with our method’s results averaged across individual user-level models. Our approach achieves competitive Acc@1 performance, and notably, the GP-F model conditioned on the causally sensitive feature Smin outperforms others in terms of MRR.

\begin{table}[!htb]
\caption{GeoLife: Average prediction performance of next location across users.}
\label{tab:combined_Results}
\centering
\resizebox{\columnwidth}{!}{
\begin{tabular}{lcccc}
\hline
\textbf{Method} & \textbf{Acc@1} & \textbf{Acc@5} & \textbf{Acc@10} & \textbf{MRR} \\
\hline
LSTM & 28.4 & 55.8 & 59.1 & 19.3 \\
LSTM + Attention & 29.8 & 54.6 & 58.2 & 21.3 \\
DeepMove & 26.1 & 54.2 & 58.7 & 38.2 \\
MHSA & 31.4 & 56.4 & 60.8 & 42.5 \\
\textbf{Proposed GCSP} \\$F_{CS}$ =Smin & 31.9 & 59.2 & 64.2 & 43.9 \\
\hline
\end{tabular}}
\end{table}
%\subsection{Ablation Study} 
\item \textbf{Ablation Study:}
%We perform an ablation study by conditioning on the causally sensitive feature and subsequently removing it as presented in Table \ref{tab:ablation_study}. In the Baseline scenario, during the training of generative predictor, we do not consider any causally sensitive features for conditioning. The results demonstrate a significant performance improvement when causally sensitive features are used for conditioning. Specifically, for the case of Smin, the accuracy  \textbf{Acc@1:}  is increased by 10.34 $\%$, and the mean reciprocal rank (MRR) improved by 5.00$\%$ . For W,  the accuracy Acc@1 is increased by 3.77 $\%$ and MRR remains almost the same. In contrast, there was no improvement for DS, as it does not have a significant causal influence. Additionally, we notice that the impact of altered location sequence LS2 is moderately higher than that of altered location sequence LS1 on LS (original location sequence).

We perform an ablation study by conditioning on the causally sensitive feature and subsequently removing it, as presented in Table~\ref{tab:ablation_study}. In the Baseline scenario, the generative predictor is trained without conditioning on any causally sensitive features. The results demonstrate significant performance improvement when such features are used for conditioning. Specifically, conditioning on Smin increases \textbf{Acc@1} by 10.34$\%$ and improves the mean reciprocal rank (MRR) by 5.00\%. For W, \textbf{Acc@1} increases by 3.77\%, with MRR remaining nearly unchanged. In contrast, no improvement is observed for DS, confirming its minimal causal influence. Additionally, the impact of altered location sequence LS2 is moderately higher than that of LS1 on the original location sequence LS.
\end{enumerate}
\begin{table}[!h]
\caption{GeoLife: Ablation study on conditioning with causally sensitive features.}
\label{tab:ablation_study}
\resizebox{\columnwidth}{!}{
\begin{tabular}{lcccc}
\hline
\textbf{Scenario} & \textbf{Acc@1} & \textbf{Acc@5} & \textbf{Acc@10} & \textbf{MRR} \\
\hline
Baseline (No Conditioning)          & 28.895 & 56.018 & 61.737 & 41.897 \\
Conditioning on \\\textbf{Start Minute (Smin)} & 32.018 & 57.980 & 63.647 & 44.310 \\
Conditioning on \\Weekday (W)         & 29.092 & 58.002 & 62.798 & 41.636 \\
Conditioning on \\Duration of Stay (DS) & 18.859 & 44.205 & 49.755 & 32.576 \\\hline
\end{tabular}
}
\end{table}

%\begin{table}[h]
%\centering
%\resizebox{\linewidth}{!}{%
%
%\begin{tabular}{l c c c c}
%%\hline
%\textbf{Scenario} & \textbf{Acc@1} & \textbf{Acc@5} & \textbf{Acc@10} & \textbf{MRR} \\
%%\hline
%Baseline (No Conditioning) & 28.895 & 56.018 & 61.737 & 41.897 \\
%Conditioning on Start Minute (Smin) & 32.018 & 57.980 & 63.647 & 44.310 \\
%Conditioning on Weekday (W) & 29.092 & 58.002 & 62.798 & 41.636 \\
%Conditioning on Duration of Stay (DS) & 18.859 & 44.205 & 49.755 & 32.576 \\
%%\hline
%%\textbf{Performance Gain (\%)} & \textbf{+10.34} & \textbf{+1.75} & \textbf{+3.09} & \textbf{+5.00} \\
%%\hline
%\end{tabular}%
%}
%\caption{Ablation Study: Impact of Conditioning on Causally Sensitive Features}
%\label{tab:ablation_study}
%\end{table
%\begin{table}[!h]
%\caption{GeoLife: Ablation Study on Conditioning with Causally Sensitive Features}
%\label{tab:ablation_study}
%\resizebox{\columnwidth}{!}{
%\begin{tabular}{lcccc}
%\hline
%\textbf{Scenario} & \textbf{Acc@1} & \textbf{Acc@5} & \textbf{Acc@10} & \textbf{MRR} \\
%\hline
%Baseline (No Conditioning)          & 28.895 & 56.018 & 61.737 & 41.897 \\
%Conditioning on \\\textbf{Start Minute (Smin)} & 32.018 & 57.980 & 63.647 & 44.310 \\
%Conditioning on \\Weekday (W)         & 29.092 & 58.002 & 62.798 & 41.636 \\
%Conditioning on \\Duration of Stay (DS) & 18.859 & 44.205 & 49.755 & 32.576 \\
%\hline
%\end{tabular}}
%\end{table}
%
Although not detailed in this paper, we have validated the proposed method on a cross-city mobility dataset, further confirming its ability to identify causally sensitive features across diverse spatiotemporal settings.
%\textbf{The summary of key differentiators for demonstrating on human trajectory is as follows:} 
%
%\begin{itemize}
%\item We identify the causally sensitive features like day of week, start time of location visits etc., and associated cause-effect relationship. 
%
%%Causality sensitive features play a significant role in counterfactual data generation\cite{Yang2021} and also in retaining cause-effect relationships in the generated data. %\cite{SS23}
%\item The proposed method predicts the user's specific next location in a GPS-based human trajectory applying the identified causal sensitive variables. %We demonstrate the contributions of different features to the individual's next location prediction using factual analysis using GPS-based human trajectory. 
%\item Furthermore, we expand our causal sensitivity identification framework to identify causal sensitivity from another perspective, where we demonstrate the impact of the counterfactuals of the cause on the effect, and at the same time illustrate the alternative sequence of past location visits and how it impacts predicting the next location visit.
%\end {itemize}
\section{Discussion and Conclusion}
We have presented a novel generative causal sensitivity identification method that combines intervention and counterfactual analysis to identify causal influence in prediction tasks. %We first validate our approach using the Asia network from the Bayesian Network Repository, a benchmark with known causal structure, to confirm that our method accurately identifies causal paths and sensitive variables. We further demonstrate its practical usefulness by applying it to next-location prediction in spatiotemporal human trajectory data.

The proposed method comprises two causal perspectives. The first is to identify causally sensitive features ($F_{CS}$) through interventional analysis, reducing confounding bias by blocking backdoor paths and establishing direct causal links between cause and effect when $F_{CS}$ acts as a confounder.
The identified $F_{CS}$ are used as conditioning inputs in the CVAE-based generative predictor to obtain causally sensitive recommendation with improved factual prediction performance. 

The second perspective is to assess the change in effect when the cause has changed using counterfactual analysis to identify the causal path, and determine the counterfactual predictions in alternate situations. %We have shown the effectiveness of applying proposed causal sensitivity identification in next location prediction in spatiotemporal human trajectory comprising timeseries of GPS data.

We validate our approach using the Asia Bayesian network benchmark. This dataset allows us to verify whether the proposed method can uncover known causal relationships under controlled conditions. We demonstrate that interventions on variables like, \textit{either}, \textit{bronc} lead to significant changes in prediction of downstream nodes such as \textit{dysp}, confirming the method's ability to identify true causal paths. Counterfactual evaluations further highlight the impact of modifying key variables, showing divergence in prediction behavior consistent with the known causal structure.
Additionally, our method outperforms CausalVAE in counterfactual sensitivity analysis for the Asia dataset, more accurately identifying direct and indirect causal influences on the target variable.

Applied to the GeoLife GPS trajectory dataset, our method identifies day of the week and start time as causally sensitive features influencing both past and next locations, while duration of stay shows minimal impact. Counterfactual sensitivity is assessed by altering past visits, which reveals shifts in predictions and divergence in latent space. Conditioning on causally sensitive features yields the best performance in factual next location prediction, establishing their importance for this task. Compared to prior works using the same input structure, our method achieves competitive results.
While manual testing of individual features is possible, such empirical approaches lack guarantees of causal relevance and may reflect spurious correlations. Our method offers a unified solution by quantifying causal significance through interventions and counterfactuals. %It enhances both robustness and interpretability, surpassing the limitations of purely empirical methods.

In summary, our generative causal sensitivity identification method provides a generalizable and interpretable framework for analyzing causal relationships, particularly in prediction tasks where the causal graph is unknown and no structural constraints (such as acyclicity) are imposed during learning. This approach is not limited to human mobility and is extendable to a wide range of applications involving time-series prediction, classification, and personalized recommendation, offering the potential for both performance gains and causal interpretability.

\bibliographystyle{named}
%\bibliography{causal_sensitivity}

\begin{thebibliography}{}

\bibitem[\protect\citeauthoryear{Bandyopadhyay and Sarkar}{2023}]{SBS}
S.~Bandyopadhyay and S.~Sarkar.
\newblock Exploring causality aware data synthesis.
\newblock In {\em Proc. ACM AIMLSystems}, 2023.

\bibitem[\protect\citeauthoryear{Dai \bgroup \em et al.\egroup
  }{2025}]{dai2025selection}
Haoyue Dai, Yaqi Xue, Krzysztof Chalupka, and Elias Bareinboim.
\newblock When selection meets intervention: Additional complexities in causal
  discovery.
\newblock In {\em International Conference on Learning Representations (ICLR)},
  2025.

\bibitem[\protect\citeauthoryear{Doersch}{2021}]{VAE-T}
C.~Doersch.
\newblock Tutorial on variational autoencoders.
\newblock {\em arXiv preprint arXiv:2111.10846}, 2021.

\bibitem[\protect\citeauthoryear{Feng \bgroup \em et al.\egroup
  }{2018}]{deepmove}
J.~Feng, L.~Yong, C.~Zhang, F.~Sun, F.~Meng, A.~Guo, and D.~Jin.
\newblock Deepmove: Predicting human mobility with attentional recurrent
  networks.
\newblock In {\em Proc. WWW Conf.}, pages 1459--1468, 2018.

\bibitem[\protect\citeauthoryear{Feng \bgroup \em et al.\egroup
  }{2020}]{movesim}
J.~Feng, Z.~Yang, F.~Xu, H.~Yu, M.~Wang, and Y.~Li.
\newblock Learning to simulate human mobility.
\newblock In {\em Proceedings of the 26th ACM SIGKDD Conference on Knowledge
  Discovery and Data Mining}, pages 3426--3433, 2020.

\bibitem[\protect\citeauthoryear{Goodfellow \bgroup \em et al.\egroup
  }{2014}]{goodfellow2014generative}
I.~J. Goodfellow, J.~Pouget-Abadie, M.~Mirza, B.~Xu, D.~Warde-Farley, S.~Ozair,
  A.~Courville, and Y.~Bengio.
\newblock Generative adversarial nets.
\newblock In {\em Advances in Neural Information Processing Systems}, 2014.

\bibitem[\protect\citeauthoryear{Hershey and Olsen}{2007}]{KlD}
John~R. Hershey and Peder~A. Olsen.
\newblock Approximating the kullback--leibler divergence between gaussian
  mixture models.
\newblock In {\em Proceedings of the IEEE International Conference on
  Acoustics, Speech and Signal Processing (ICASSP)}, volume~4, pages IV--317.
  IEEE, 2007.

\bibitem[\protect\citeauthoryear{Hong \bgroup \em et al.\egroup
  }{2023}]{ref_article1}
Y.~Hong, Y.~Zhang, K.~Schindler, and M.~Raubal.
\newblock Context-aware multi-head self-attentional neural network model for
  next location prediction.
\newblock {\em Transp. Res. Part C: Emerg. Technol.}, 156, 2023.

\bibitem[\protect\citeauthoryear{Ke \bgroup \em et al.\egroup
  }{2020}]{ke2020learning}
Nan~Rosemary Ke, Olexa Bilaniuk, Anirudh Goyal, Stephan Bauer, Hugo Larochelle,
  Chris Pal, and Yoshua Bengio.
\newblock Learning neural causal models from unknown interventions.
\newblock In {\em International Conference on Learning Representations (ICLR)},
  2020.

\bibitem[\protect\citeauthoryear{Kingma and Welling}{2013}]{kingma2019auto}
D.~P. Kingma and M.~Welling.
\newblock Auto-encoding variational bayes.
\newblock {\em arXiv preprint arXiv:1312.6114}, 2013.

\bibitem[\protect\citeauthoryear{Kment}{2020}]{countercausal}
Boris Kment.
\newblock Counterfactuals and causal reasoning.
\newblock In {\em Perspectives on Causation: Selected Papers from the Jerusalem
  2017 Workshop}, pages 463--482. Springer, 2020.

\bibitem[\protect\citeauthoryear{Krishna \bgroup \em et al.\egroup
  }{2018}]{krishna2018}
Kalpit Krishna, Devendra Jain, Shobhit~V. Mehta, and Shubham Choudhary.
\newblock An lstm-based system for prediction of human activities with
  durations.
\newblock {\em Proceedings of the ACM on Interactive, Mobile, Wearable and
  Ubiquitous Technologies}, 1(4):147:1--147:31, 2018.

\bibitem[\protect\citeauthoryear{Lauritzen and
  Spiegelhalter}{1988}]{lauritzen1988local}
Steffen~L. Lauritzen and David~J. Spiegelhalter.
\newblock Local computations with probabilities on graphical structures and
  their application to expert systems.
\newblock {\em Journal of the Royal Statistical Society: Series B
  (Methodological)}, 50(2):157--194, 1988.

\bibitem[\protect\citeauthoryear{Li \bgroup \em et al.\egroup }{2019}]{klAnn}
Chunyuan Li, Xiujun Liu, Jianfeng Gao, Asli Celikyilmaz, and Lawrence Carin.
\newblock Cyclical annealing schedule: A simple approach to mitigating kl
  vanishing.
\newblock In {\em Proceedings of the 2019 Conference of the North {A}merican
  Chapter of the Association for Computational Linguistics: Human Language
  Technologies (NAACL-HLT)}, pages 240--250. Association for Computational
  Linguistics, 2019.

\bibitem[\protect\citeauthoryear{Li \bgroup \em et al.\egroup }{2020}]{li2020}
Fei Li, Zhen Gui, Zhili Zhang, Dawei Peng, Shuang Tian, Kai Yuan, Yafei Sun,
  Huayi Wu, Jing Gong, and Yinjie Lei.
\newblock A hierarchical temporal attention-based lstm encoder-decoder model
  for individual mobility prediction.
\newblock {\em Neurocomputing}, 403:153--166, 2020.

\bibitem[\protect\citeauthoryear{Liu and others}{2021}]{trajgan}
X.~Liu et~al.
\newblock Trajgans: Geo-privacy protection of trajectory data.
\newblock In {\em GIScience}, 2021.

\bibitem[\protect\citeauthoryear{Louizos \bgroup \em et al.\egroup
  }{2017}]{CEVAE}
Christos Louizos, Uri Shalit, Joris Mooij, David Sontag, Rich Zemel, and Max
  Welling.
\newblock Causal effect inference with deep latent-variable models.
\newblock In {\em Proceedings of the Advances in Neural Information Processing
  Systems (NeurIPS)}, 2017.

\bibitem[\protect\citeauthoryear{MacDonald and Zucchini}{1997}]{hmm}
Iain~L. MacDonald and Walter Zucchini.
\newblock {\em Hidden Markov and Other Models for Discrete-Valued Time Series},
  volume 110 of {\em Monographs on Statistics and Applied Probability}.
\newblock CRC Press, Boca Raton, FL, 1997.

\bibitem[\protect\citeauthoryear{Martin \bgroup \em et al.\egroup
  }{2022}]{trackintel}
H.~Martin, Y.~Hong, N.~Wiedemann, D.~Bucher, and R.~Martin.
\newblock Trackintel: An open-source python library for human mobility
  analysis.
\newblock {\em arXiv preprint arXiv:2206.03593}, 2022.

\bibitem[\protect\citeauthoryear{Nezhadettehad \bgroup \em et al.\egroup
  }{2024}]{nezhadettehad2024}
Amin Nezhadettehad, Arkady Zaslavsky, Rafiq Abdur, S.~A. Shaikh, Seng~W. Loke,
  Guang-Li Huang, and Ali Hassani.
\newblock Predicting next useful location with context-awareness: The
  state-of-the-art.
\newblock {\em arXiv preprint arXiv:2401.08081}, 2024.

\bibitem[\protect\citeauthoryear{Pearl and Mackenzie}{2018}]{bookofwhy}
Judea Pearl and Dana Mackenzie.
\newblock {\em The Book of Why: The New Science of Cause and Effect}.
\newblock Basic Books, 2018.

\bibitem[\protect\citeauthoryear{Pearl}{2009}]{Pearl-C}
J.~Pearl.
\newblock {\em Causality: Models, reasoning, and inference}.
\newblock Cambridge University Press, New York, 2nd edition, 2009.

\bibitem[\protect\citeauthoryear{Pearl}{2019}]{Pearl-Counter}
J.~Pearl.
\newblock {\em Causal and counterfactual inference}.
\newblock Springer, New York, 2019.

\bibitem[\protect\citeauthoryear{Rudenko \bgroup \em et al.\egroup
  }{2020}]{Rudenko2020}
Andrey Rudenko, Luigi Palmieri, Michael Herman, Kris~M. Kitani, Dariu~M.
  Gavrila, and Kai~O. Arras.
\newblock Human motion trajectory prediction: A survey.
\newblock {\em International Journal of Robotics Research}, 39(8):895--935,
  2020.

\bibitem[\protect\citeauthoryear{Salzmann \bgroup \em et al.\egroup
  }{2020}]{salzmann2020trajectron}
T.~Salzmann, B.~Ivanovic, P.~Chakravarty, and M.~Pavone.
\newblock Trajectron++: Multi-agent generative trajectory forecasting with
  heterogeneous data for control.
\newblock {\em CoRR}, abs/2001.03093, 2020.

\bibitem[\protect\citeauthoryear{Scutari}{2009}]{bnlearn}
Marco Scutari.
\newblock The bnlearn dataset repository.
\newblock \url{https://www.bnlearn.com/bnrepository/}, 2009.
\newblock Accessed: 2025-05-25.

\bibitem[\protect\citeauthoryear{van~der Maaten and Hinton}{2008}]{tsne}
Laurens van~der Maaten and Geoffrey Hinton.
\newblock Visualizing data using t-sne.
\newblock {\em Journal of Machine Learning Research}, 9:2579--2605, 2008.

\bibitem[\protect\citeauthoryear{Vaswani \bgroup \em et al.\egroup
  }{2017}]{attention}
A.~Vaswani, N.~Shazeer, N.~Parmar, J.~Uszkoreit, L.~Jones, A.~N. Gomez,
  L.~Kaiser, and I.~Polosukhin.
\newblock Attention is all you need.
\newblock In {\em Proc. 31st NeurIPS}, pages 5998--6008. Curran Associates,
  2017.

\bibitem[\protect\citeauthoryear{Wang \bgroup \em et al.\egroup
  }{2022}]{Wang2022}
S.~Wang, J.~Cao, and P.~S. Yu.
\newblock Deep learning for spatio-temporal data mining: A survey.
\newblock {\em IEEE Trans. Knowl. Data Eng.}, 34(8):3681--3700, 2022.

\bibitem[\protect\citeauthoryear{Xia \bgroup \em et al.\egroup }{2023}]{count1}
Kun Xia, Yang Pan, and Elias Bareinboim.
\newblock Causal models for counterfactual identification and estimation.
\newblock In {\em Proceedings of the International Conference on Learning
  Representations (ICLR)}, 2023.

\bibitem[\protect\citeauthoryear{Yang \bgroup \em et al.\egroup
  }{2018}]{Yang2018}
Karren Yang, Abigail Katcoff, and Caroline Uhler.
\newblock Characterizing and learning equivalence classes of causal dags under
  interventions.
\newblock In {\em Proceedings of the 35th International Conference on Machine
  Learning (ICML)}, pages 5541--5550. PMLR, 2018.

\bibitem[\protect\citeauthoryear{Yang \bgroup \em et al.\egroup
  }{2021a}]{Yang2021}
Mengyang Yang, Fanqian Liu, Zhiqian Chen, Xinyu Shen, Jingyao Hao, and Jun
  Wang.
\newblock Causalvae: Disentangled representation learning via neural structural
  causal models.
\newblock In {\em Proceedings of the IEEE Conference on Computer Vision and
  Pattern Recognition (CVPR)}, pages 9593--9602, 2021.

\bibitem[\protect\citeauthoryear{Yang \bgroup \em et al.\egroup
  }{2021b}]{Yang2021CausalVAE}
Mengyang Yang, Feng Liu, Zhihui Chen, Xiaohui Shen, Jun Hao, and Jingdong Wang.
\newblock Causalvae: Structured causal disentanglement in variational
  autoencoder.
\newblock In {\em Proceedings of the IEEE/CVF Conference on Computer Vision and
  Pattern Recognition (CVPR)}, pages 9593--9602. IEEE, 2021.

\bibitem[\protect\citeauthoryear{Yang \bgroup \em et al.\egroup
  }{2021c}]{causalattention}
Xinrui Yang, Hongyang Zhang, Guosheng Qi, and Jianbo Cai.
\newblock Causal attention for vision-language tasks.
\newblock In {\em Proceedings of the IEEE Conference on Computer Vision and
  Pattern Recognition (CVPR)}, 2021.

\bibitem[\protect\citeauthoryear{Zheng \bgroup \em et al.\egroup
  }{2010}]{geolife}
Y.~Zheng, X.~Xie, and W.-Y. Ma.
\newblock Geolife: A collaborative social networking service among user,
  location and trajectory.
\newblock {\em IEEE Data Eng. Bull.}, 33(2):32--39, 2010.

\bibitem[\protect\citeauthoryear{Zheng \bgroup \em et al.\egroup
  }{2018}]{NOTEARS}
Xun Zheng, Bryon Aragam, Pradeep Ravikumar, and Eric~P. Xing.
\newblock Dags with no tears: Continuous optimization for structure learning.
\newblock In {\em Advances in Neural Information Processing Systems (NeurIPS)},
  volume~31, 2018.

\bibitem[\protect\citeauthoryear{Zuo \bgroup \em et al.\egroup
  }{2022}]{count-kung}
A.~Zuo, S.~Wei, T.~Liu, B.~Han, K.~Zhang, and M.~Gong.
\newblock Counterfactual fairness with partially known causal graph.
\newblock In {\em Proc. AAAI Conference on Artificial Intelligence}, 2022.

\end{thebibliography}

\end{document}